\documentclass{article}

\usepackage[preprint, nonatbib]{nips_2018}

\usepackage[utf8]{inputenc} 
\usepackage[T1]{fontenc}    
\usepackage{hyperref}       
\usepackage{url}            
\usepackage{booktabs}       
\usepackage{amsfonts}       
\usepackage{nicefrac}       
\usepackage{microtype}      

\usepackage{verbatim}       
\usepackage{subcaption}
\usepackage{graphicx}
\usepackage{float}
\usepackage{color,soul}
\usepackage{adjustbox}
\usepackage{amsmath}
\usepackage{bm}
\usepackage{mathtools}
\usepackage{algorithmic}
\usepackage{algorithm}
\usepackage{placeins}
\usepackage{tikz}
\usepackage{dsfont}
\usepackage{enumitem}
\usetikzlibrary{positioning}
\usepackage{multirow}
\usepackage[flushleft]{threeparttable}

\mathchardef\mhyphen="2D 
\DeclarePairedDelimiterX{\infdiv}[2]{(}{)}{%
	#1\;\delimsize\|\;#2%
}

\begin{document}

\title{Adversarial Attacks on Variational Autoencoders}

\author{
   George Gondim-Ribeiro, Pedro Tabacof, and Eduardo Valle \\
   RECOD Lab. — DCA / School of Electrical and Computer Engineering (FEEC)\\
   University of Campinas (Unicamp)\\
   Campinas, SP, Brazil \\
   \texttt{ \{gribeiro, tabacof, dovalle\}@dca.fee.unicamp.br } \\
}

\maketitle

\begin{abstract}
Adversarial attacks are malicious inputs that derail machine-learning models. We propose a scheme to attack autoencoders, as well as a quantitative evaluation framework that correlates well with the qualitative assessment of the attacks. We assess --- with statistically validated experiments --- the resistance to attacks of three variational autoencoders (simple, convolutional, and DRAW) in three datasets (MNIST, SVHN, CelebA), showing that both DRAW's recurrence and attention mechanism lead to better resistance. As autoencoders are proposed for compressing data --- a scenario in which their safety is paramount --- we expect more attention will be given to adversarial attacks on them.
\end{abstract}

\section{Introduction}

Adversarial attacks derail models by crafting malicious inputs. Image classifiers mislabel those inputs --- visually indistinguishable from ordinary ones --- with high confidence.

\begin{figure}[b]
    \centering
    \includegraphics[width=0.8\textwidth]{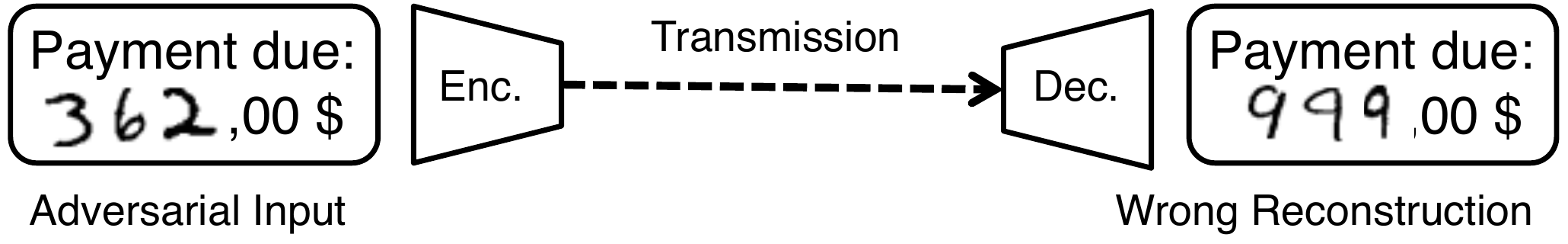}
    \caption{A motivating example: an attacker may convince a sender to transmit a seemingly innocent document that is reconstructed in a malicious way. (Scenario inspired by~\cite{kos2017advexam}).}
    \label{fig:motivation}
\end{figure}

In comparison to the extensive literature on adversarial attacks for classifiers, attacks for autoencoders are mostly unexplored, possibly because those attacks are hard both to perform and to assess. Still, as autoencoders are advanced as powerful schemes for compressing information~\cite{gregor2016towards}, attacks on them are potentially at least as dangerous as attacks on classifiers. In a hypothetical example, an attacker may present someone a document (e.g., a contract, or agreement), which upon transmission turns into a different one (Figure ~\ref{fig:motivation}).

Evaluating generative models is hard~\cite{theis2015note}, there are no clear-cut success criteria for autoencoder reconstruction, and therefore, neither for the attack. We bypass that difficulty by analyzing how inputs and outputs differ across varying compromises between distorting the input and approaching the target.

In this paper, we propose a scheme to attack autoencoders, as well as a quantitative evaluation framework for the attacks that bypass the need for a success criterion. We compare three kinds of autoencoders: simple variational autoencoders (with fully-connected layers), convolutional variational autoencoders, and DRAW --- a recently proposed recurrent variational autoencoder with an attention mechanism~\cite{gregor2015draw}. We show that the latter is more resistant to the attacks, and that its recurrent and attention mechanism both contribute to the resistance. We run all --- statistically validated --- experiments in three datasets (MNIST, SVHN, and CelebA) and show that our quantitative assessment correlates well with a qualitative perception of the attacks.

\section{Autoencoders, Generative Models, and Variational Autoencoders}

\begin{figure}[ht]
    \centering
    \includegraphics[width=0.65\textwidth]{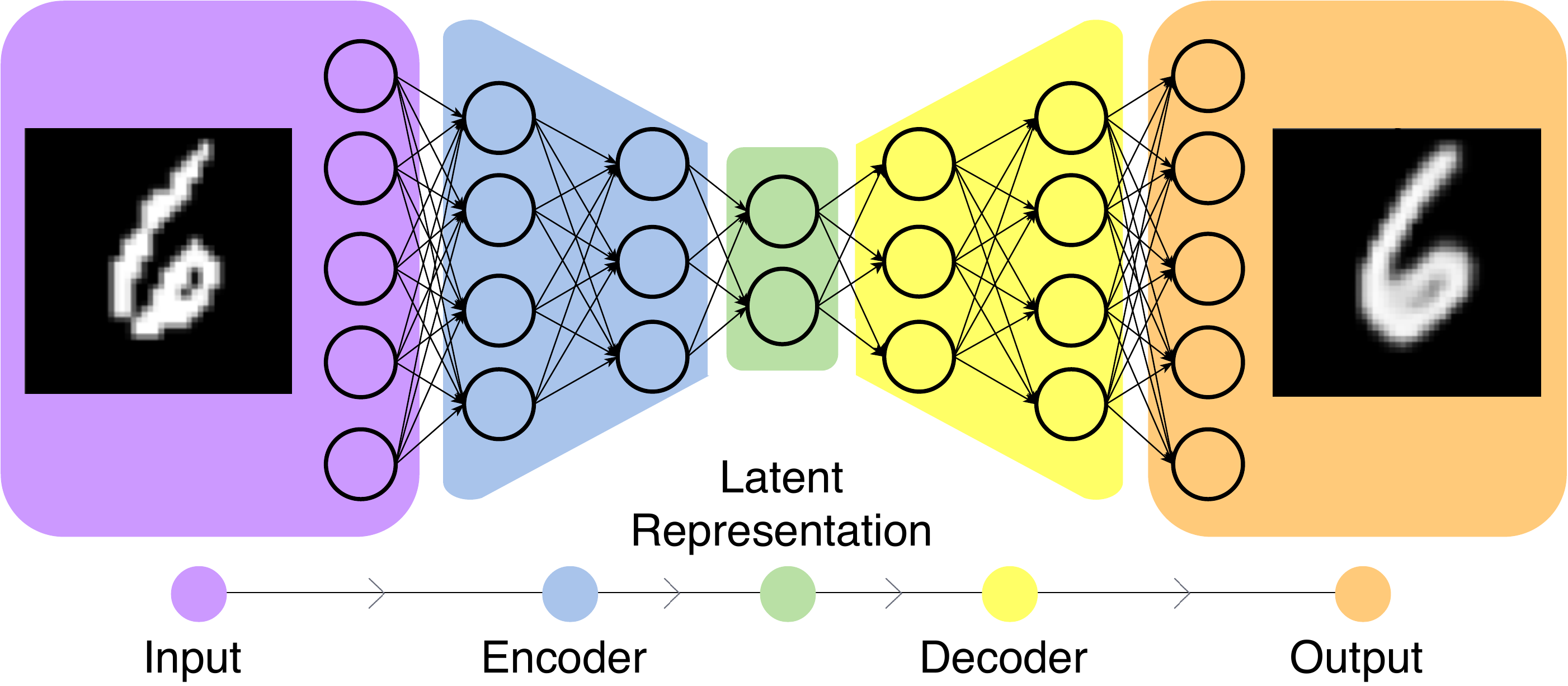}
    \caption{Autoencoders are models that map their input into a (deterministic or stochastic) latent representation, and then map such representation into an output similar to the input; those two maps form the two halves of the model: the encoder and the decoder.}
    \label{fig:autoencoderDiagram}
\end{figure}

Autoencoders are neural networks that compress their input into a smaller latent representation (encoder), and then reconstruct the input from that representation (decoder). Autoencoders are trained end-to-end by backpropagation, with a loss that reflects the difference between input and output, e.g., $\ell_2$. (Figure ~\ref{fig:autoencoderDiagram})

Autoencoders admit many variations: sparse~\cite{ng2011sparse}, denoising~\cite{vincent2010stacked}, variational~\cite{kingma2013auto, rezende2014stochastic}, Wasserstein~\cite{tolstikhin2017wasserstein}, symmetric~\cite{pu2017advsymmetric}, etc. They differ on the encoding/decoding architectures, on the computation of the loss, or on the noise inserted into the network.

Variational autoencoders (VAEs) are particularly interesting since they behave as both autoencoders and as generative models. VAEs have stochastic latent representations, which, once trained, can be decoded to generate samples similar (but not identical) to those in the original dataset. Together with Generative Adversarial Networks (GANs)~\cite{goodfellow2014generative}, VAEs are the most studied deep generative models. There are attempts to unify VAEs and GANs into a single framework~\cite{pu2017advsymmetric, larsen2015vaegan, makhzani2016advvae, mescheder2017advvaebayes, makhazani2017pixel}. Although that generative aspect of VAEs is one of their main appeals, in this paper we will approach them only as autoencoders.

In VAEs, the latent representation is a parametric distribution. The encoder outputs the parameters of that distribution from the observed input. The decoder samples the latent distribution and tries to reconstruct something similar to the input. VAEs learn the encoder and decoder parameters end-to-end by backpropagation, maximizing the probability that sampling from the latent distribution leads to a good reconstruction of the input, in a maximum marginal likelihood framework.

The encoder and decoder can be implemented with different neural architectures: multilayer perceptron~\cite{kingma2013auto}, convolutional~\cite{radford2015unsupervised}, or even recurrent~\cite{hochreiter1997long}. We are particularly interested in the recently proposed DRAW~\cite{gregor2015draw}, which uses LSTMs with an attention mechanism, using a metaphor of ``painting'' the image in a canvas, step by step.

Formally, VAEs are probabilistic models that aim at maximizing $p_\theta(\bm{x}), x\in X$, where $X$ is our training set, and $\theta$ are the network parameters. We introduce a stochastic latent representation $z$ and maximize the marginal probability distribution~(Equation~\ref{eq:problatent}) over it:

\begin{equation}
\label{eq:problatent}
p_\theta(\bm{x}) = \int p_\theta(\bm{x},\bm{z}) d\bm{z} = \int p_\theta(\bm{x}|\bm{z})p(\bm{z})d\bm{z}
\end{equation} 

In theory, the prior $p(\bm{z})$ is arbitrary, but in practice a standard multivariate Gaussian $\mathcal{N}(0,I)$ is most often employed~\cite{kingma2013auto}. Since the integration in Equation~\ref{eq:problatent} is intractable, we opt to maximize its variational lower bound (Equation~\ref{eq:elbo}), where $\phi\subset\theta$ are the encoder parameters and $\mathrm{D_{KL}}$ is the KL divergence.

\begin{equation}
\label{eq:elbo}
\begin{aligned}
\log p_\theta(\bm{x}) \ge
\mathrm{E}_{q_{\phi}(\bm{z}|\bm{x})}[\log p_\theta(\bm{x}|\bm{z})] - \mathrm{D_{KL}}\infdiv{q_\phi (\bm{z}|\bm{x})}{p(\bm{z})}
=-\mathrm{D_{KL}}\infdiv{q_\phi (\bm{z}|\bm{x})}{p(\bm{z}|\bm{x})}
\end{aligned}
\end{equation} 

In variational approaches, an approximate posterior is used in place of the true posterior $p_{\phi}(\bm{z}|\bm{x})$. In VAEs, this approximate posterior is $\mathrm{decoder}_\psi(q_{\phi}(\bm{z}|\bm{x}))$, where $\psi\subset\theta$ are the parameters of the decoder and $q_{\phi}(\bm{z}|\bm{x})$ usually is an uncorrelated multivariate Gaussian as shown in Equation~\ref{eq:encoder}. 

\begin{equation}
\label{eq:encoder}
q_\phi(\bm{z}|\bm{x}) = \mathcal{N}(\bm{\mu}_{\phi}(\bm{x}), \exp(\bm{\sigma}_{\phi}^2(\bm{x})))
\end{equation} 

VAEs exploit the universal approximator property of neural networks~\cite{hornik1991approximation} to use a Gaussian distribution in Equation~\ref{eq:elbo}, and transform the posterior by learning the decoder parameters. That approach makes the problem tractable, since the KL divergence now has analytic form~\cite{kingma2013auto}. Finally, the ``reparameterization trick'' allows the model to be differentiated end-to-end~\cite{kingma2015variational}. The trick consists on parameterizing the random variables ${\bm z} \sim q_{\phi}(\bm{z}|\bm{x})$ as a differentiable transformation of a noise variable ${\bm \epsilon} \sim \mathcal{N}(0, I)$ such as ${\bm z} \sim \bm{\mu}_{\phi}(\bm{x}) + \bm{\sigma}_{\phi}({\bm x}){\bm \epsilon}$.

\section{Adversarial Attacks on Variational Autoencoders} 

Following the seminal paper of Szegedy et al.~\cite{szegedy2013intriguing}, adversarial attacks on neural networks classifiers attracted much attention~\cite{goodfellow2014explaining, tabacof2015exploring, kurakin2016adversarial}. Those attacks aim at creating small distortions on the input (most often images) that lead to misclassification. Attacks can aim at a target wrong class~\cite{szegedy2013intriguing, tabacof2015exploring, dezfooli2016deepfool,carlini2016towards} (targeted), or they can aim at any class other than the right one (untargeted)~\cite{goodfellow2014explaining, kurakin2016advexamples}. State-of-the-art attacks produce essentially imperceptible distortions that make the classifier predict the wrong class with high confidence.

\begin{figure}[ht]
    \centering
    \includegraphics[width=0.75\textwidth]{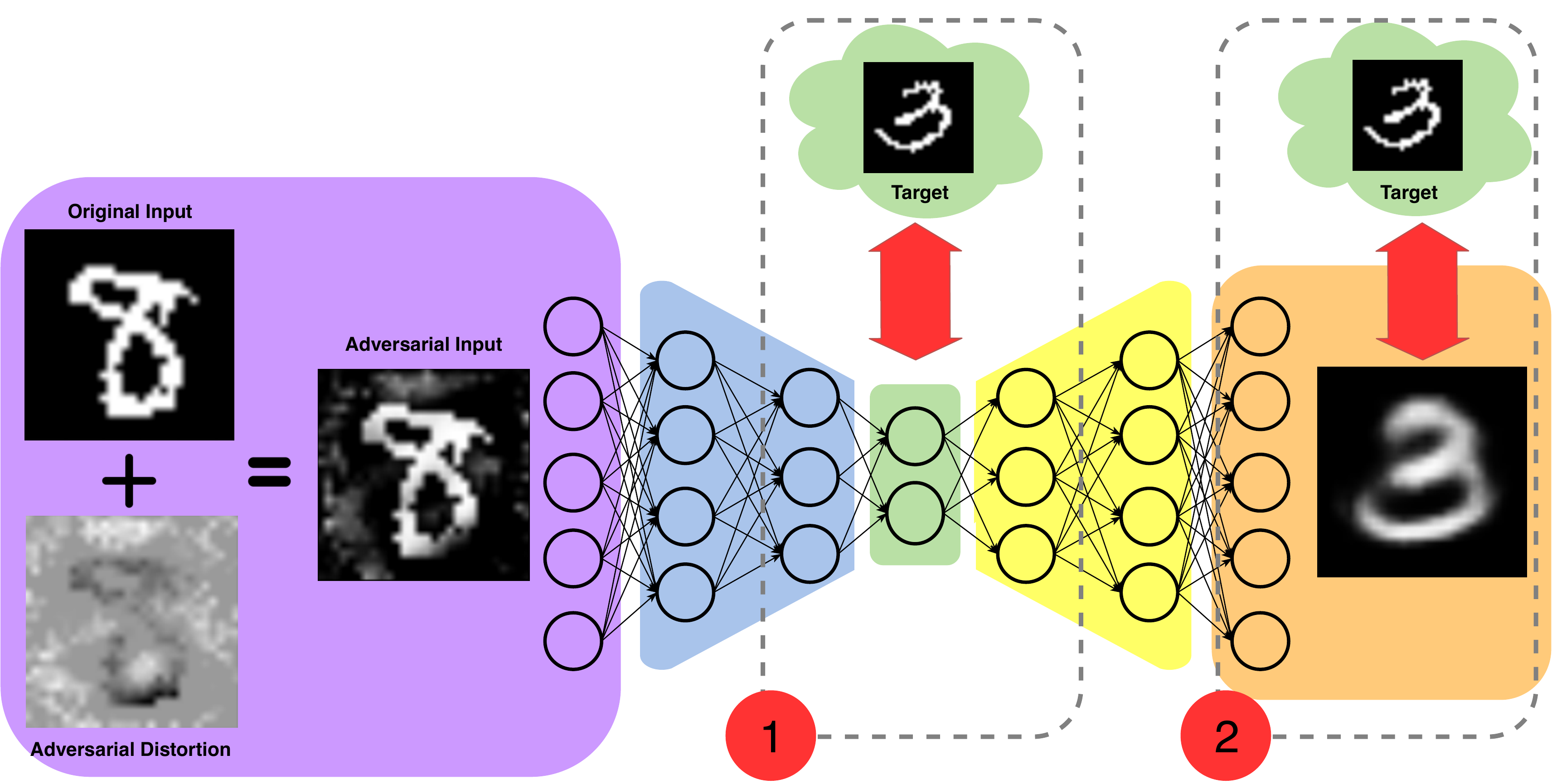}
    \caption{Adversarial attacks for autoencoders aim at disrupting the reconstruction as much as possible, while minimizing the distortion on the input. On targeted attacks, the challenge is inducing the autoencoder to reconstruct a different sample. We evaluate attacks on the latent representation layer (1), and on the output layer (2).}
    \label{fig:attackOnAutoencoder}
\end{figure}

Attacking autoencoders follows a parallel course: the aim is to induce minimal distortions on the input that disrupt the reconstructed output (Figure~\ref{fig:attackOnAutoencoder}). The attack can aim at reconstructing a particular wrong input (targeted), or just at thwarting the reconstruction (untargeted). In this paper we focus on targeted attacks. 

Compared to attacks on classifiers, attacks on autoencoders are much less explored. Sabour et al.~\cite{sabour2015adversarial}, while still working on classifiers, introduced the notion of attacking internal layers of deep neural networks. Tabacof et al.~\cite{tabacof2016advvae} introduced attacks on autoencoders and variational autoencoders, showing that they are possible, although much harder than attacks on classifiers. They attacked the latent representation with a KL-divergence objective in both MNIST and SVHN. They proposed the graphs we call Distortion--Distortion plots here and evaluated attack success by visual inspection of those graphs. They also showed that there is a linear compromise between the intensity of the input distortion and the degree of success in the attack for both autoencoders and classifiers. Kos et al.~\cite{kos2017advexam} followed up with a work that attacked both the latent representation and the output of VAE--GAN autoencoders. They proposed three modes of attack: attacking an extraneous classifier after the latent representation, attacking the latent representation directly with an $\ell_2$ objective, and attacking the output of the decoder using the VAE loss function. They introduced a quantitative, although indirect, evaluation of attack inferred from success in fooling the extraneous classifier.

Attacking autoencoders is a more involved procedure than attacking classifiers. In the latter we target a small output vector, often focusing at just one or two values on that vector. In the former we need to address a very high-dimensional output. Targeted attacks to autoencoders consist in adding (as small as possible) adversarial distortion to the original input in order make the reconstructed output as close as possible to the target (Figure~\ref{fig:attackOnAutoencoder}).

Attacks are performed on trained models, whose weights are kept constant, while minimizing the change to the input. The most ``obvious'' attack also minimizes the difference between the output and the target, as explained in Equation~\ref{eq:ae_adv_out}.

\begin{table}[h]
\begin{equation}
\label{eq:ae_adv_out}
\noindent
  \thinmuskip=\muexpr\thinmuskip*5/8\relax
  \medmuskip=\muexpr\medmuskip*5/8\relax
\begin{aligned}
& \underset{\bm{d}}{\min}
& & \Delta(\bm{r_{a}}, \bm{I_{t}}) + C\| \bm{d} \| \\
& \text{s.t.}
& & L \leq \bm{x} + \bm{d} \leq U ,\quad
\bm{z_{a}} = \mathrm{encoder}(\bm{x} + \bm{d}) ,\quad
\bm{r_{a}} = \mathrm{decoder}(\bm{z_a})
\end{aligned}
\end{equation} 
\end{table}

where $\bm{d}$ is the adversarial distortion, $\bm{x}+\bm{d}$ is the adversarial input, and its output reconstruction $\bm{r_a}$ is reconstructed from a sample of $\bm{z_{a}}$ (the latent representation, which in variational autoencoders is a distribution). $L$ and $U$ are the bounds of the input space, i.e., $L\leq \bm{x} \leq U, \forall \bm{x}$ that is valid as input to the $\mathrm{encoder}$. $C$ is the regularization constant that balances approaching the target and limiting the input distortion. $\bm{I_t}$ is the target, and $\Delta$ is the distance used to compare it to the output, in our case, $\ell_2$.

A less obvious attack minimizes the difference between \textit{latent representations}, which attacks the network at its smallest layer. That attack implies solving the optimization in Equation~\ref{eq:ae_adv}. 

\begin{table}[h]
\begin{equation}
\label{eq:ae_adv}
\noindent
  \thinmuskip=\muexpr\thinmuskip*5/8\relax
  \medmuskip=\muexpr\medmuskip*5/8\relax
\begin{aligned}
& \underset{\bm{d}}{\min}
& & \Delta(\bm{z_{a}}, \bm{z_{t}}) + C\| \bm{d} \| \\
& \text{s.t.}
& & L \leq \bm{x} + \bm{d} \leq U ,\quad
\bm{z_{a}} = \mathrm{encoder}(\bm{x} + \bm{d})
\end{aligned}
\end{equation} 
\end{table}

where $\bm{z_{t}}$ is the latent representation of the target, and the other symbols are the same as in Equation~\ref{eq:ae_adv_out}. Here, use the KL-divergence as $\Delta$. Although it is not a true metric, it intuitively measures the (asymmetric) difference between two distributions, so it is a proxy for the distance between the latent representations.

\section{Measuring the Success of Attacks on Autoencoders}
\label{sec:metric}

When attacking classifiers, there is a clear-cut criterion for success: the target class has higher probability than all others (on targeted attacks), or the right class has lower probability than some other (on untargeted attacks). That criterion is used, in one way or another, to perform the attack: one can optimize the input distortion until that criterion is satisfied~\cite{szegedy2013intriguing, tabacof2015exploring}, establish a maximum input distortion and observe if the criterion is satisfied~\cite{goodfellow2014explaining, sabour2015adversarial},  etc. 

When attacking autoencoders, we want to maximize the disruption on the output (untargeted attacks) or its similarity to a target image (targeted attacks), while minimizing the input distortion, but there is no sharp criterion for success. 

\begin{figure}[ht!]
      \centering
      \begin{subfigure}[b]{0.48\textwidth}
              \includegraphics[width=\textwidth]{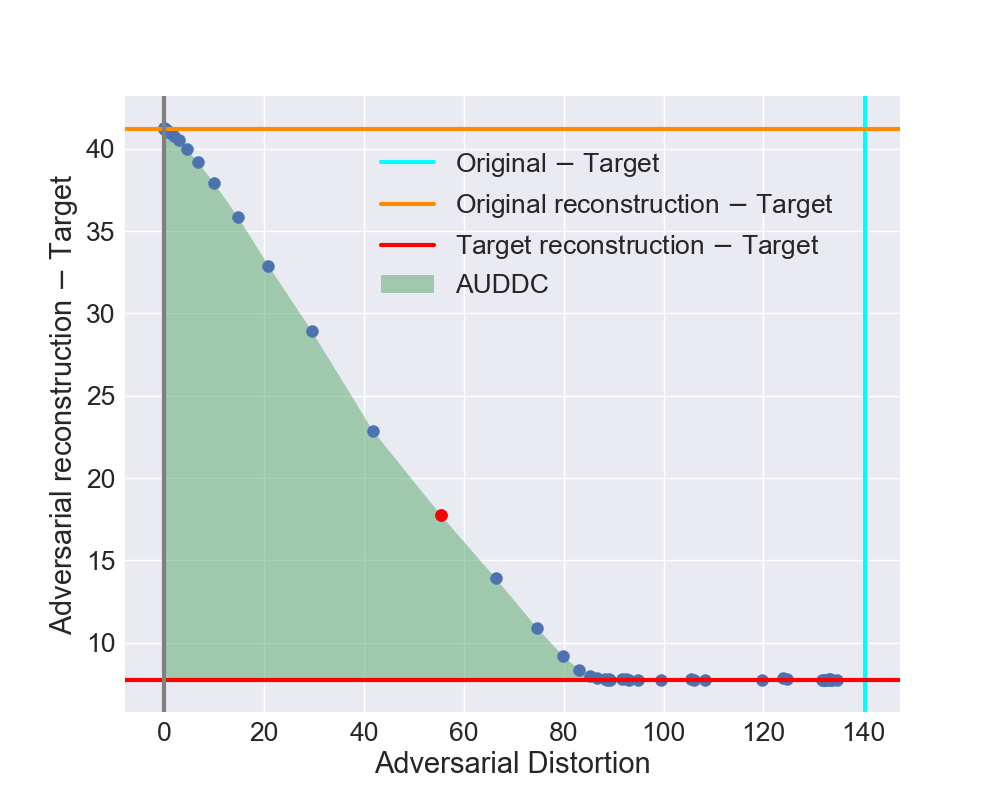}
      \end{subfigure}
      \begin{subfigure}[b]{0.48\textwidth}
          \includegraphics[width=\textwidth]{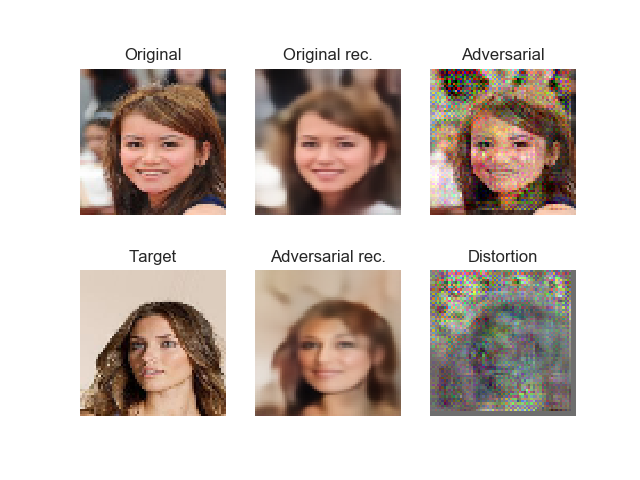}
      \end{subfigure}
      \caption{Left: the proposed metric: Area Under the Distortion--Distortion Curve (AUDDC). Right: visualization of a single point (red dot) of the left plot.}
      \label{fig:metric}
\end{figure}

For targeted attacks, Tabacof et al.~\cite{tabacof2016advvae} proposed graphs that examine the entire spectrum of compromises between approaching the target and distorting the original (Figure~\ref{fig:metric}). They do not name such graphs, which we choose to call \textbf{Distortion--Distortion plots}. Using a set of boundaries (explained below), they normalize those graphs, and use visual inspection on stacked normalized plots of various cases to compare attacks.

Kos et al.~\cite{kos2017advexam} provide a quantitative metric based on an ancillary classifier network. The classifier uses the latent representation as input, and is trained on labels that may be arbitrary, but must have a relationship to the input (they cannot be random labels). They compute two metrics: success rate ignoring the target (how often the reconstruction of the attack input mislead the classifier), and success considering the target (how often the reconstruction of the attack input matched with the class of the target). 

Those propositions have complementary drawbacks. Tabacof et al.'s allows only qualitative analysis. Kos et al.'s evaluate a different criterion than we evaluate here: not whether the target is reconstructed (in semantics \textit{and} appearance), but whether it reproduces the semantics of the class. It works only for labeled data. 

We address those shortcomings with the \textbf{AUDDC} (Area under Distortion--Distortion Curve). For a given original and target pair, we compute different results, with different approximation compromises. All attacks can vary, in one way or another, this compromise: in our case, we vary the values of $C$ in Equations~\ref{eq:ae_adv_out} and~\ref{eq:ae_adv}). The Distortion--Distortion plots show, for each attempt, how much we distorted the original and how much we approached the target (both measured by $\ell_2$). We add limiting lines to the plot: no distortion added (and original reconstruction) at the leftmost/gray and topmost/orange lines; the $\ell_2$-distance between the target and the reconstruction of the target by the model at the bottommost/red line; the $\ell_2$-distance between the original and target image. Those limits represent, respectively, the starting point, the intrinsic limitation of the model, and the maximum ``sensible'' distortion (which allows going from the original to the target directly). We normalize the graph so that the distance between those lines is 1. The AUDDC is the area under the curve given by the linear interpolation of the points. The closer this area is to 1, the more resistant the model was to the attack (the less successful the attack was).

\section{Data and Methods}

\textbf{Datasets:} We employed three datasets --- MNIST~\cite{lecun1998mnist}, SVHN~\cite{netzer2011reading}, and CelebA~\cite{liu2015faceattributes} --- with the respective training and test splits. We expected those datasets to present increasing levels of challenge for the autoencoders: MNIST has handwritten decimal single digits, without color, SVHN has multi-digit street numbers in several styles and colors, and CelebA has human faces in color.

\textbf{Models:} We evaluated four models --- variational autoencoder with fully-connected layers as encoder/decoder (VAE); variational autoencoder with convolutional layers as encoder and deconvolutional layers as decoder (CVAE); the recurrent autoencoder DRAW~\cite{gregor2015draw} without and with its attention mechanism. Following  literature~\cite{kingma2013auto}, we modeled pixel likelihoods as independent Bernoullis in MNIST, and as independent Gaussians in SVHN and CelebA. In all models, the latent representations ($\bm{z_a}$ and $\bm{z_t}$ in Equations~\ref{eq:ae_adv_out} and~\ref{eq:ae_adv}) are uncorrelated multivariate normal distributions with parameters given by the encoder.

Extended hyperparameter exploration was not the scope of this study. Still, adversarial attacks are not interesting if the attacked model is bad in itself. To obtain good reconstructions, we started with hyperparameters given by deep learning ``guild knowledge'' on each dataset, and made minimal fine-tuning until the reconstruction quality (without attack) was good. For DRAW, we start with the models suggested in the original paper~\cite{gregor2015draw}. The quality of each model can be appreciated in Figure~\ref{fig:qualitativeResults}. We give details about the models below:

\textit{VAE (all datasets)} --- Encoder and Decoder: fully(512)~$\rightarrow$ fully(512).

\textit{CVAE--MNIST} --- Encoder:~conv(32,4)~$\rightarrow$ conv(64,4)~$\rightarrow$ conv(128,4)~$\rightarrow$ fully(512); Decoder:~fully(512)~$\rightarrow$ dconv(128,3)~$\rightarrow$ dconv(64,3)~$\rightarrow$ dconv(32,2)~$\rightarrow$ dconv(16,2).

\textit{CVAE--SVHN} --- Encoder:~same as MNIST; Decoder:~fully(512)~$\rightarrow$ dconv(128,5)~$\rightarrow$ dconv(64,5)~$\rightarrow$ dconv(32,5).

\textit{CVAE--CelebA} --- Encoder:~conv(32,4)~$\rightarrow$ conv(64,4)~$\rightarrow$ conv(128,4)~$\rightarrow$ conv(256,4)~$\rightarrow$ fully(512); Decoder:~fully(512)~$\rightarrow$ dconv(256,5)~$\rightarrow$ dconv(128,5)~$\rightarrow$ dconv(64,5)~$\rightarrow$ dconv(32,5).

\textit{DRAW--MNIST} --- Encoder and Decoder: LSTM~= 256 units, attention window~= 8$\times$8 pixels.

\textit{DRAW--SVHN} --- Encoder and Decoder: LSTM~= 256 units, attention window~= 16$\times$16 pixels.

\textit{DRAW--CelebA} --- Encoder and Decoder: LSTM~= 2500 units (1 timestep)~/ 400 units (16 timesteps), attention window~= 24$\times$24 pixels.

In the notation above, fully($x$) is a fully-connected layer with $x$ units, and (d)conv($x$,$y$) is a (de)convolutional layer with $x$ filters of size $y$. All (de)convolutional layers had stride~= 2.

\textbf{Training:} Each model was trained for 500 epochs. At every 10 epochs, we evaluated the loss on the validation set -- 20\% of random samples removed from the training set -- keeping the weights that offered minimal loss. The training/validation loss was the ELBO (Equation~\ref{eq:elbo}). We approximated the expectation by sampling the posterior once. We extracted the gradients with automatic differentiation and maximized the ELBO with Adam~\cite{kingma2014adam}, with a learning rate of $10^{-4}$ and batch size of $128$.

\textbf{Evaluating:} For each dataset, we picked at random an \textit{evaluation set} with 20 original--target image pairs from the test set. The evaluation set is the same across all models evaluated, to reduce spurious variability. We show an example of a pair of input and reconstructed image in Figure~\ref{fig:autoencoderDiagram}.

To evaluate a given image-pair, we perform the attack as explained below, for 51 values of the regularization constant $C$ ($\{0\} \cup \{2^{i}\}$, with 50 values of $i$ equally spaced between $-20$ and $20$, inclusive. After all 51 attacks are completed, we compute the metric (explained in Section~\ref{sec:metric}) for that image-pair.

\textbf{Attacking:} To compute a single point on the Distortion--Distortion plot we pick a value for the regularization constant $C$ and solve the optimization problem described in Equation~\ref{eq:ae_adv} (if attacking the latent representation) or in Equation~\ref{eq:ae_adv_out} (if attacking the output). The model itself does not change during the attack: its weights remain constant.

We use L-BFGS-B~\cite{zhu1997lbfgsb} as optimizer, with initial disturbance sampled from a uniform distribution $\mathcal{U}(10^{-8}, 10^{8})$, 25 corrections on the memory matrix, and termination test tolerance of 10. All other parameters are SciPy~\cite{scipy} defaults. L-BFGS-B ensures the constraints on pixel limits, and is often employed for adversarial attacks~\cite{szegedy2013intriguing}.

Since VAEs are inherently stochastic, we performed 128 attacks to compute a single point --- we implemented this as working on a batch of images, where all inputs are the original image, and all outputs are the target image. The measured distortions for the point (Section~\ref{sec:metric}) are the average distortion for the 128 attempts.

\textbf{Experimental Design:} we considered five factors (with respective levels) in our design: (1) Dataset: MNIST, SVHN, CelebA; (2) Model: VAE, C-VAE, DRAW, DRAW-Attention; (3) Size of latent representation: Small, Large (32 \textit{vs.} 128 for MNIST and SVHN; 256 \textit{vs.} 2048 for CelebA); (4) Timesteps: 1, 16 (only for DRAW and DRAW-Attention); (5) Layer attacked: Latent, Output.

That design resulted in 72 
treatments. Each treatment is evaluated across the 20 image-pairs in the evaluation set of its dataset, and assigned the average of the evaluation metric over those pairs. 

We did both a quantitative and a qualitative analysis. For the quantitative analysis, we averaged the AUDDC for the chosen factors. To check which factors lead to significant influence, we used a multi-way ANOVA, with second-order interactions, and post-hoc Tukey honest significant differences.

For the qualitative analysis, we visually analyzed the results of five randomly selected image-pairs from each dataset.

\textbf{Hardware/Software:} We coded~\footnote{The code can be found at \url{https://github.com/gondimribeiro/adv-attacks-vae/}.} all models in Python 3.6, SciPy 1.0.0 and Tensorflow 1.4.1~\cite{tensorflow2015-whitepaper}. We ran the experiments in NVIDIA GPUs (GTX Titan X Maxwell, Titan X Pascal, and NVIDIA Tesla P100). We ran the statistical analyses in R 3.4.3. The source code, statistical scripts, and detailed results for each image-pair are available as supplemental material.

\section{Results and Analysis}

\begin{table}
  \centering
  \begin{threeparttable}
    \caption{Average $\pm$ 95\%-confidence interval of AUDDC for all models and datasets. Higher values indicate higher resistance to the attacks.}
    \label{tab:quantitativeResults}
    \begin{tabular}{l c c c c c c l}
      \toprule
              &  VAE  &  CVAE  &  DRAW*  &  DRAW  &  DRAW* &  DRAW  & \\
      Steps   &  ---  &  ---  &  1  &  1  &  16  &  16  & \\
      \midrule
              &  \multicolumn{6}{c}{Attacks on latent representation} \\
      \midrule
      MNIST   &  27 $\pm$ ~2  &  35 $\pm$ ~3  &  27 $\pm$ ~1  &  35 $\pm$ ~3  &  71 $\pm$ ~5  &  \textbf{91 $\pm$ ~3}  &  47 $\pm$ ~3 \\
      SVHN    &  19 $\pm$ ~1  &  18 $\pm$ ~1  &  09 $\pm$ ~1  &  27 $\pm$ ~2  &  74 $\pm$ ~6  &  \textbf{96 $\pm$ ~2}  &  41 $\pm$ ~4 \\
      CelebA  &  31 $\pm$ ~1  &  28 $\pm$ ~1  &  21 $\pm$ ~2  &  36 $\pm$ ~1  &  81 $\pm$ ~4  &  \textbf{97 $\pm$ ~1}  &  49 $\pm$ ~4 \\
              &  25 $\pm$ ~1  &  27 $\pm$ ~2  &  19 $\pm$ ~2  &  33 $\pm$ ~1  &  75 $\pm$ ~3  &  \textbf{95 $\pm$ ~1}  &  46 $\pm$ ~2 \\
      \midrule
              &  \multicolumn{6}{c}{Attacks on output} \\
      \midrule
      MNIST   &  35 $\pm$ ~2  &  56 $\pm$ ~3  &  38 $\pm$ ~2  &  48 $\pm$ ~4  &  29 $\pm$ ~3  &  \textbf{69 $\pm$ ~4}  &  46 $\pm$ ~2 \\
      SVHN    &  19 $\pm$ ~1  &  19 $\pm$ ~2  &  13 $\pm$ ~1  &  27 $\pm$ ~2  &  21 $\pm$ ~2  &  \textbf{34 $\pm$ ~2}  &  22 $\pm$ ~1 \\
      CelebA  &  27 $\pm$ ~1  &  24 $\pm$ ~1  &  31 $\pm$ ~3  &  35 $\pm$ ~1  &  29 $\pm$ ~2  &  \textbf{40 $\pm$ ~1}  &  31 $\pm$ ~1 \\
              &  27 $\pm$ ~1  &  33 $\pm$ ~3  &  27 $\pm$ ~2  &  37 $\pm$ ~2  &  26 $\pm$ ~1  &  \textbf{47 $\pm$ ~3}  &  33 $\pm$ ~1 \\
      \midrule
              &  \multicolumn{6}{c}{All attacks} \\
      \midrule
      MNIST   &  31 $\pm$ ~2  &  45 $\pm$ ~3  &  32 $\pm$ ~2  &  42 $\pm$ ~3  &  50 $\pm$ ~5  &  \textbf{80 $\pm$ ~3}  &  47 $\pm$ ~2 \\
      SVHN    &  19 $\pm$ ~1  &  19 $\pm$ ~1  &  11 $\pm$ ~1  &  27 $\pm$ ~1  &  47 $\pm$ ~7  &  \textbf{65 $\pm$ ~7}  &  31 $\pm$ ~2 \\
      CelebA  &  29 $\pm$ ~1  &  26 $\pm$ ~1  &  26 $\pm$ ~2  &  36 $\pm$ ~1  &  55 $\pm$ ~6  &  \textbf{68 $\pm$ ~7}  &  40 $\pm$ ~2 \\
              &  26 $\pm$ ~1  &  30 $\pm$ ~2  &  23 $\pm$ ~1  &  35 $\pm$ ~1  &  51 $\pm$ ~4  &  \textbf{71 $\pm$ ~3}  &  39 $\pm$ ~1 \\
      \bottomrule
    \end{tabular}
    \begin{tablenotes}
        \small
        \item * Attention mechanism disabled.
    \end{tablenotes}
  \end{threeparttable}
\end{table}

Table~\ref{tab:quantitativeResults} summarizes the main quantitative results. We show the averages and 95\%-confidence intervals of the AUDDC for every combination of model, dataset, number of timesteps (for DRAW), and layer attacked, as well as the marginal statistics. We averaged over the size of latent representation and the image-pairs. The values appear $\times$100 to reduce visual clutter. The ANOVA + post-hoc Tukey found significant differences (all p-values<0.015) for all pairs of levels of all factors shown on the table.

Attacking auto-encoders is relatively difficult if compared to attacking classifiers, where the distortions can be invisible to the human eye. Different models pose different challenges for the attack. DRAW, in particular, was much more resistant to our attacks --- and both its recurrent mechanism and its attention mechanism were important in conferring that resistance. The choice of datasets also influenced the challenges, with SVHN being the easiest to attack, and CelebA being the hardest. 

\begin{figure}[ht!]
      \centering
      
      \begin{subfigure}[b]{0.0525\textwidth}
          \includegraphics[width=\textwidth]{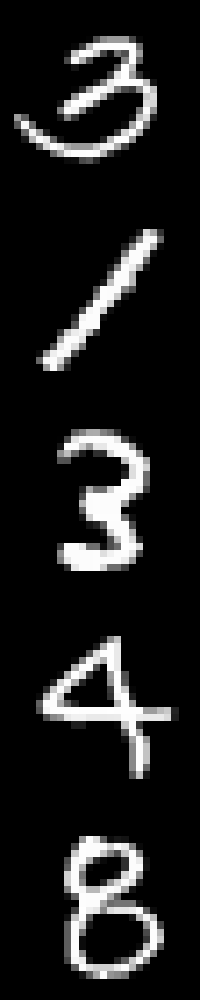}
      \end{subfigure}
      \begin{subfigure}[b]{0.21\textwidth}
          \includegraphics[width=\textwidth]{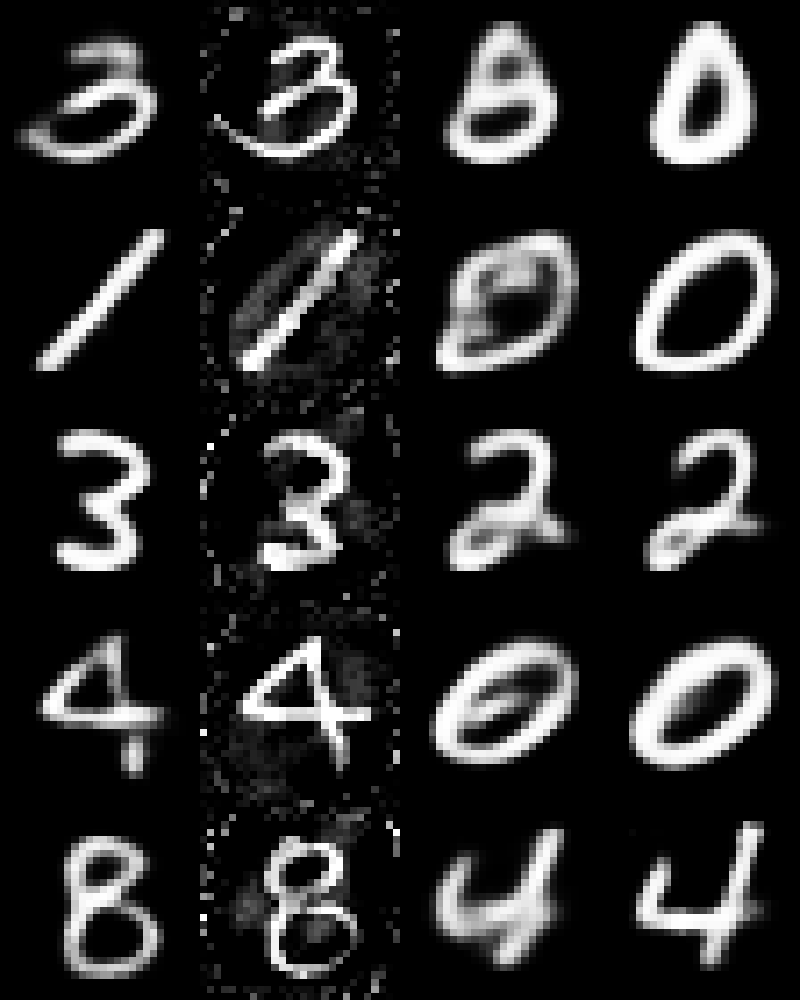}
      \end{subfigure}
      \begin{subfigure}[b]{0.21\textwidth}
          \includegraphics[width=\textwidth]{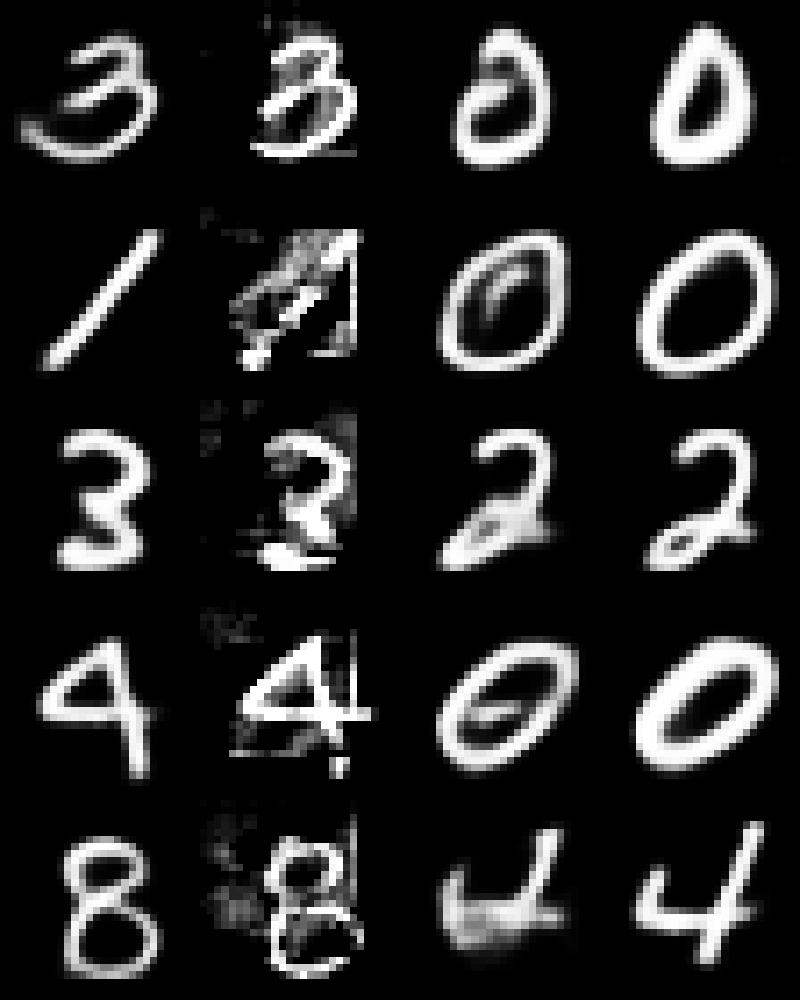}
      \end{subfigure}
      \begin{subfigure}[b]{0.21\textwidth}
          \includegraphics[width=\textwidth]{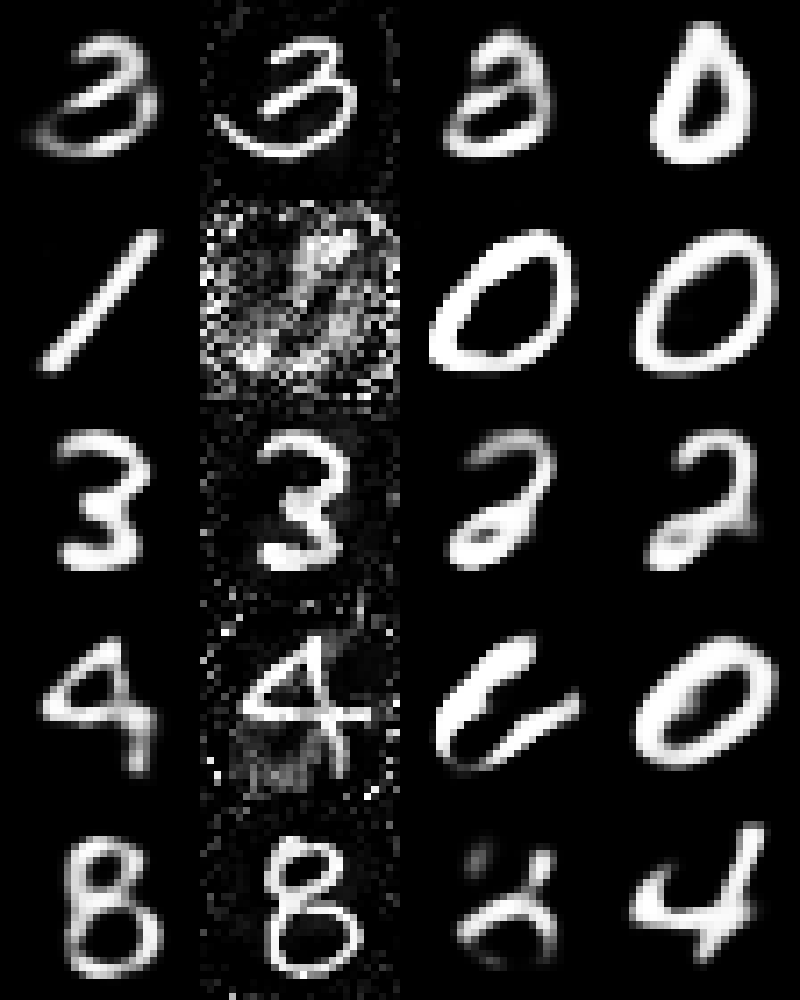}
      \end{subfigure}
      \begin{subfigure}[b]{0.21\textwidth}
          \includegraphics[width=\textwidth]{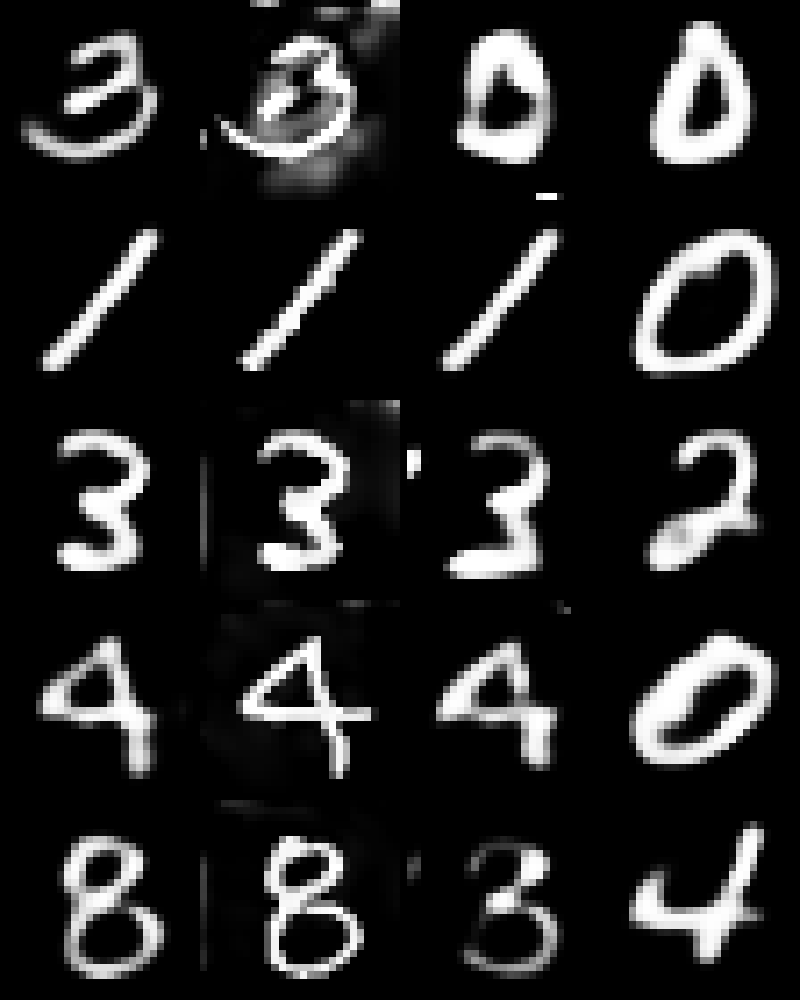}
      \end{subfigure}
      \begin{subfigure}[b]{0.0525\textwidth}
          \includegraphics[width=\textwidth]{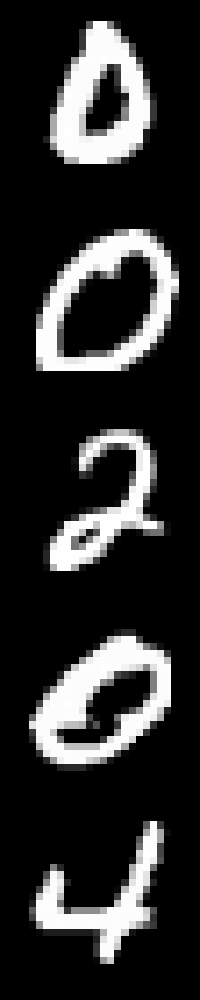}
      \end{subfigure}
      
      \begin{subfigure}[b]{0.0525\textwidth}
          \includegraphics[width=\textwidth]{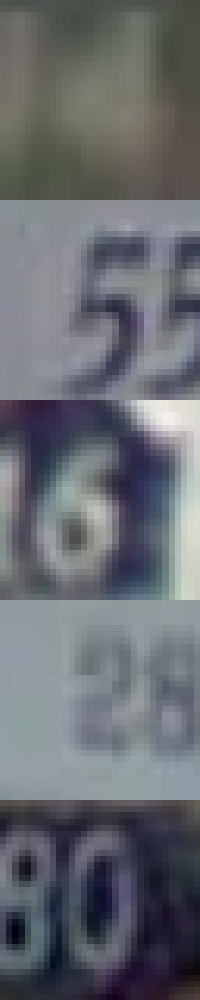}
      \end{subfigure}
      \begin{subfigure}[b]{0.21\textwidth}
          \includegraphics[width=\textwidth]{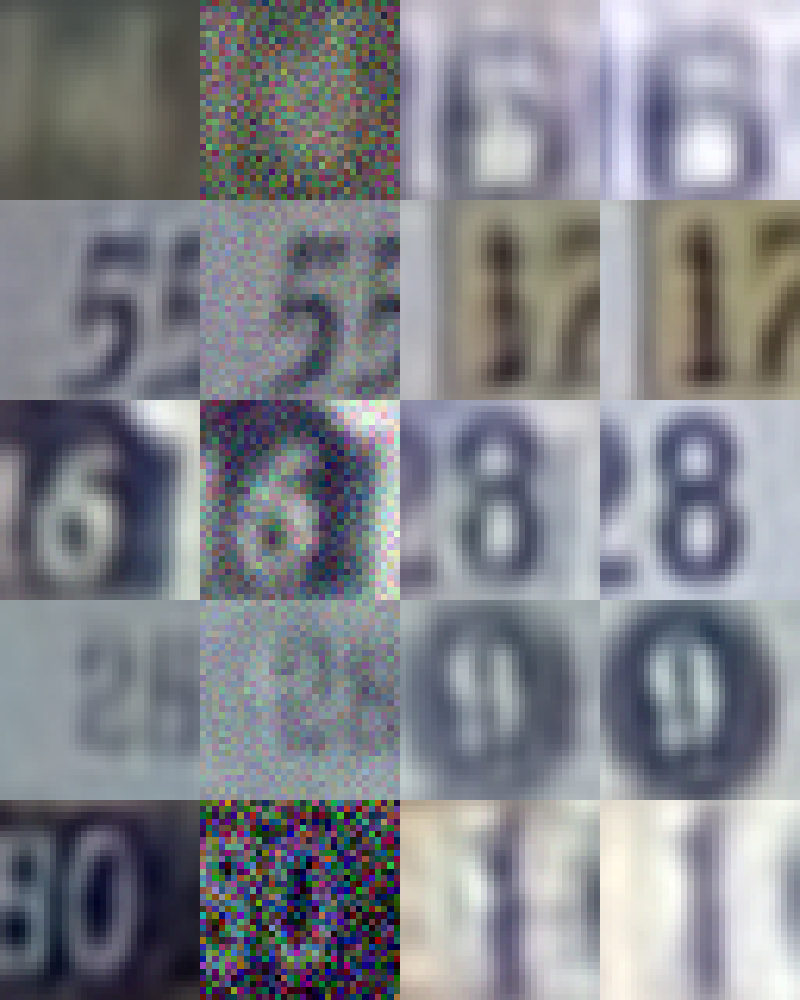}
      \end{subfigure}
      \begin{subfigure}[b]{0.21\textwidth}
          \includegraphics[width=\textwidth]{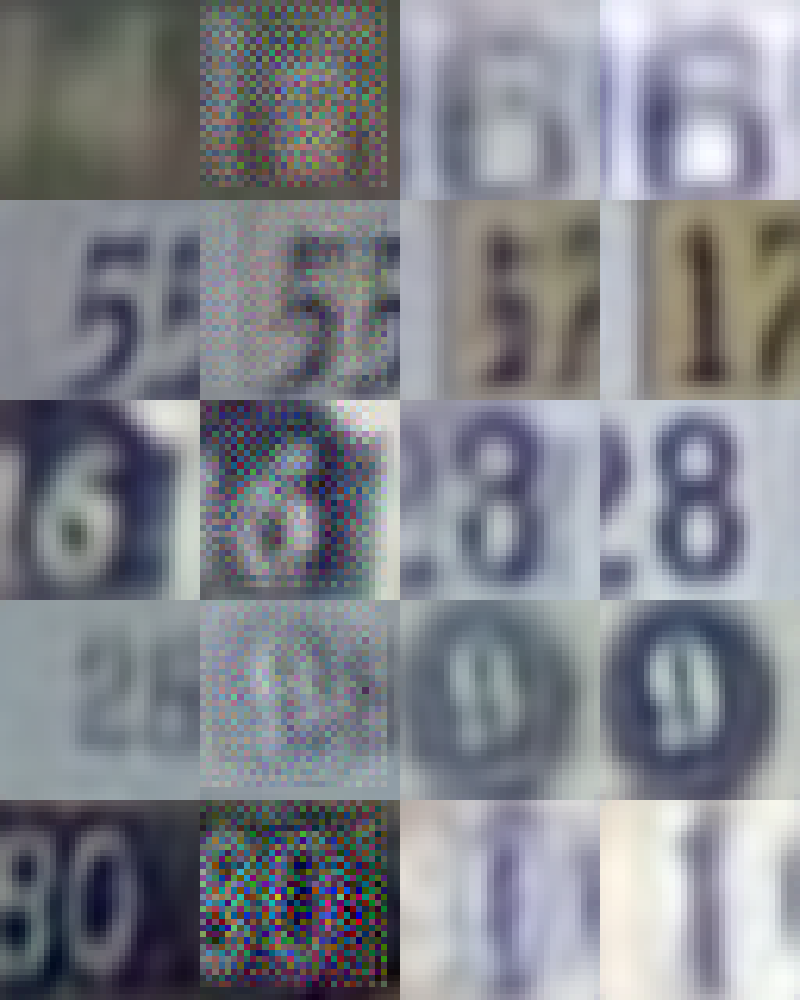}
      \end{subfigure}
      \begin{subfigure}[b]{0.21\textwidth}
          \includegraphics[width=\textwidth]{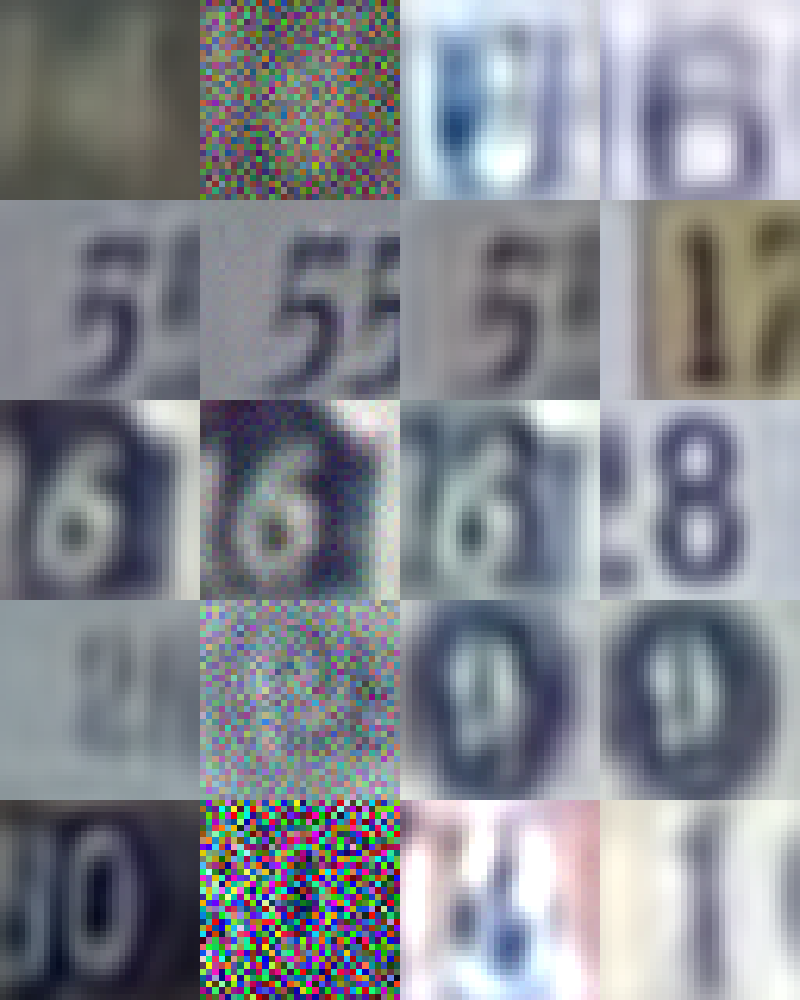}
      \end{subfigure}
      \begin{subfigure}[b]{0.21\textwidth}
          \includegraphics[width=\textwidth]{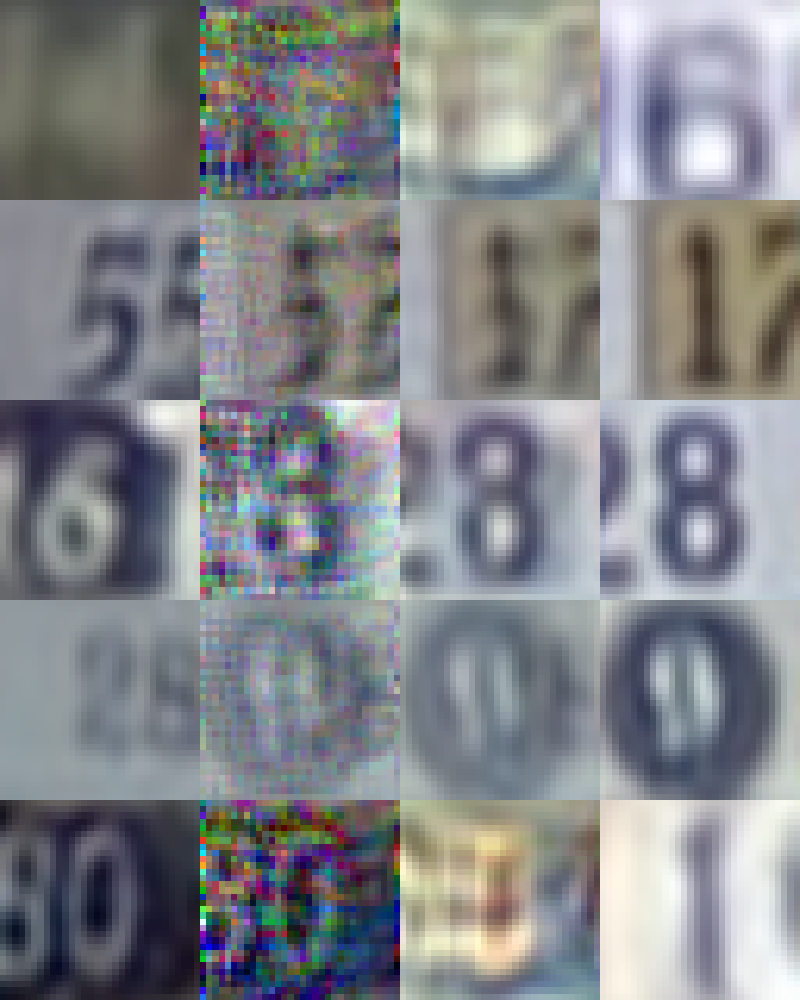}
      \end{subfigure}
      \begin{subfigure}[b]{0.0525\textwidth}
          \includegraphics[width=\textwidth]{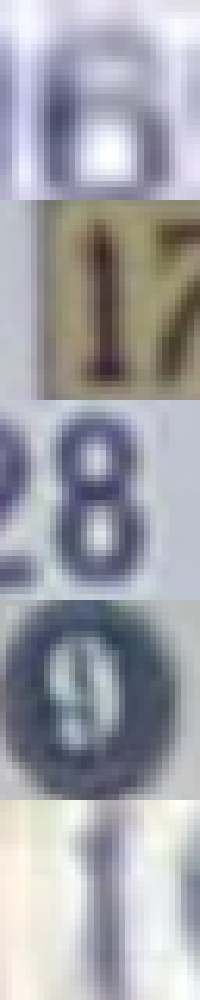}
      \end{subfigure}
      
      \begin{subfigure}[b]{0.0525\textwidth}
          \includegraphics[width=\textwidth]{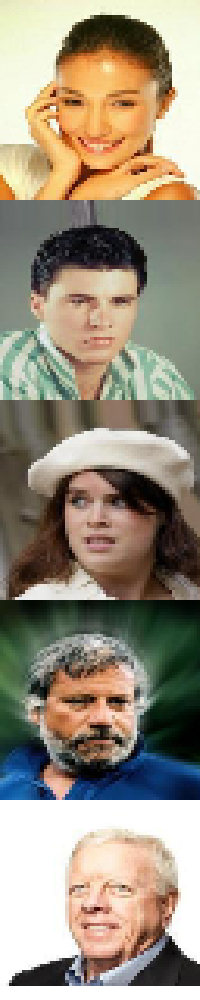}
          \caption{}
      \end{subfigure}
      \begin{subfigure}[b]{0.21\textwidth}
          \includegraphics[width=\textwidth]{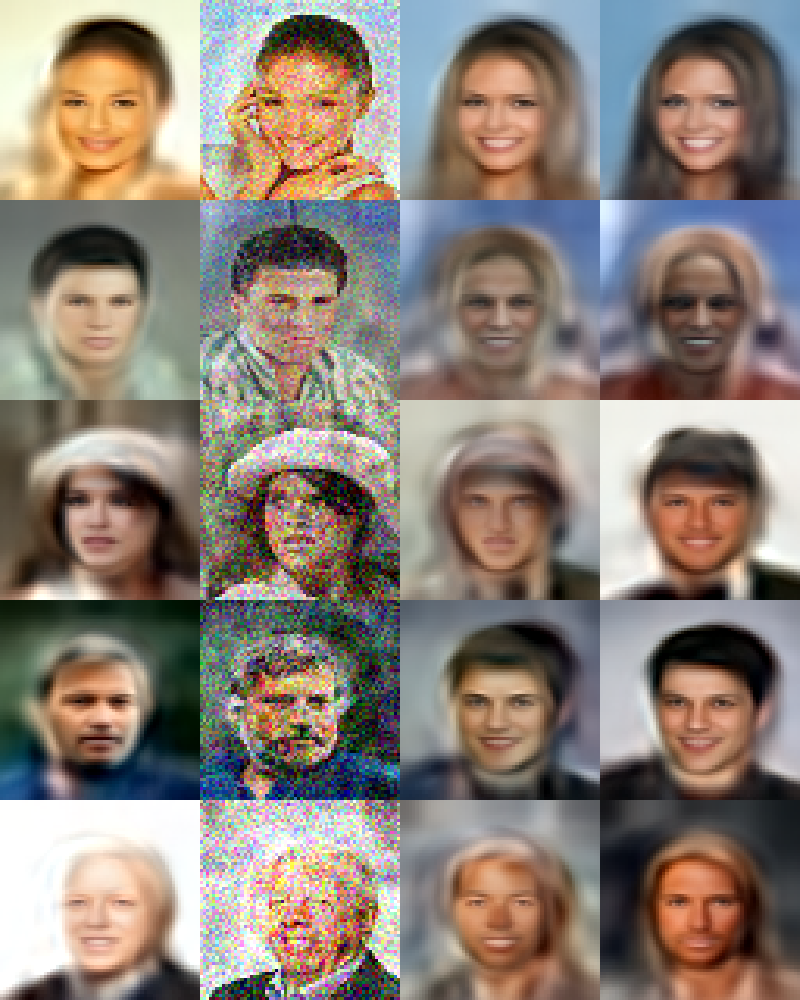}
          \caption{}
      \end{subfigure}
      \begin{subfigure}[b]{0.21\textwidth}
          \includegraphics[width=\textwidth]{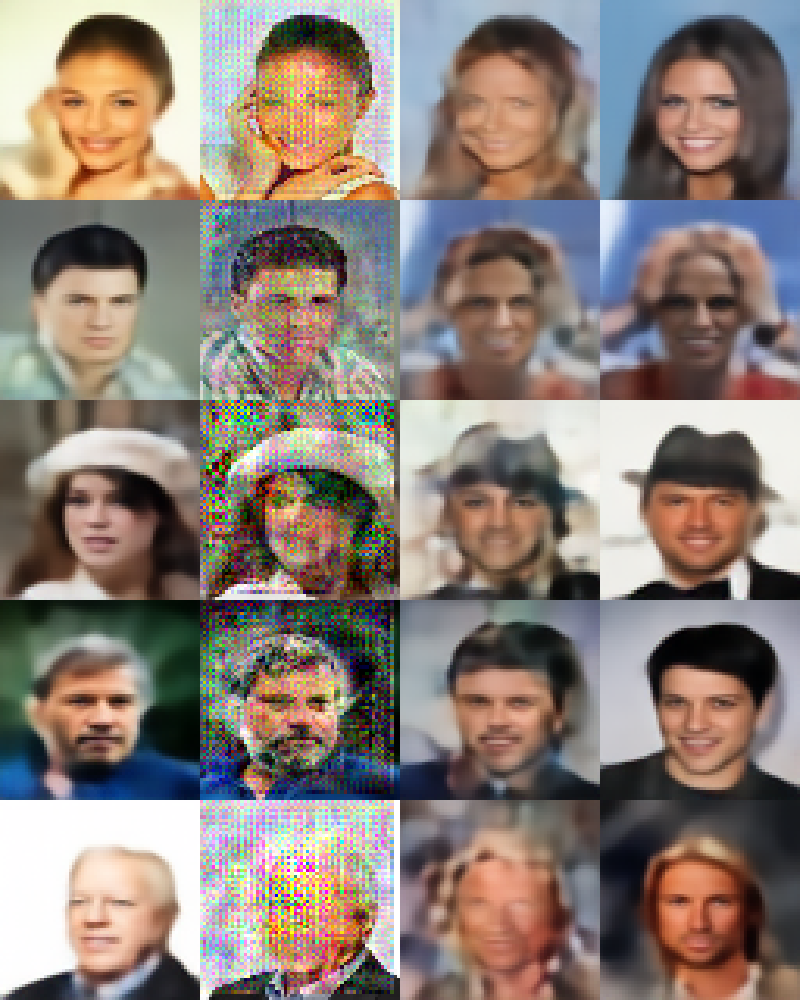}
          \caption{}
      \end{subfigure}
      \begin{subfigure}[b]{0.21\textwidth}
          \includegraphics[width=\textwidth]{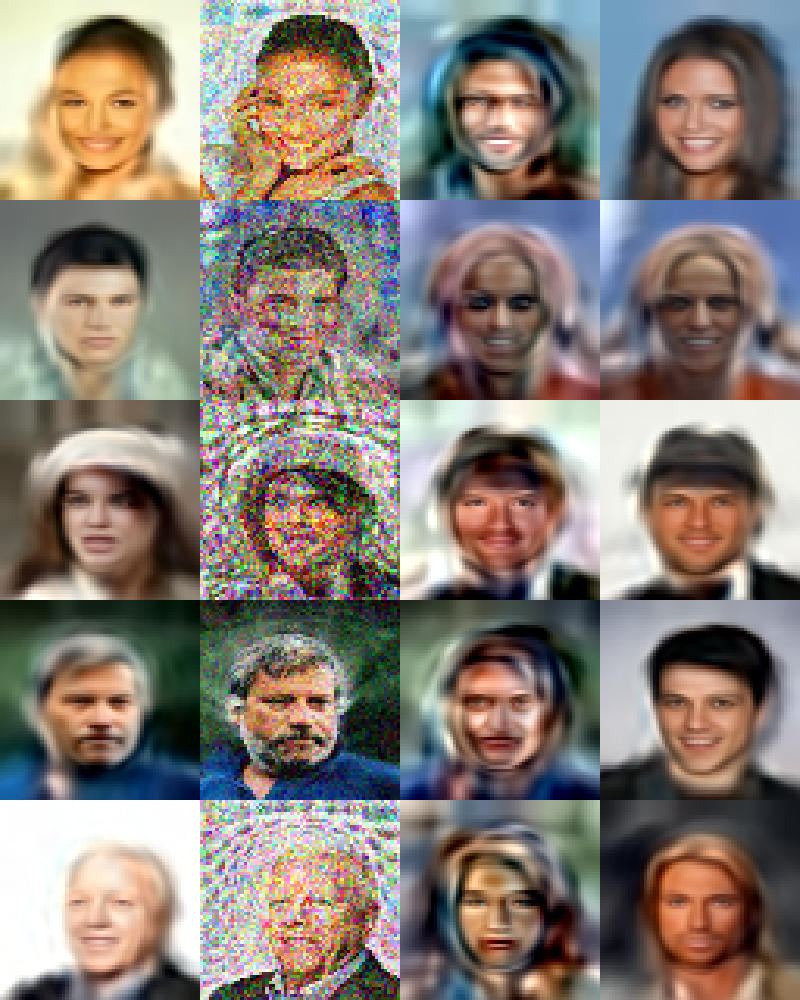}
          \caption{}
      \end{subfigure}
      \begin{subfigure}[b]{0.21\textwidth}
          \includegraphics[width=\textwidth]{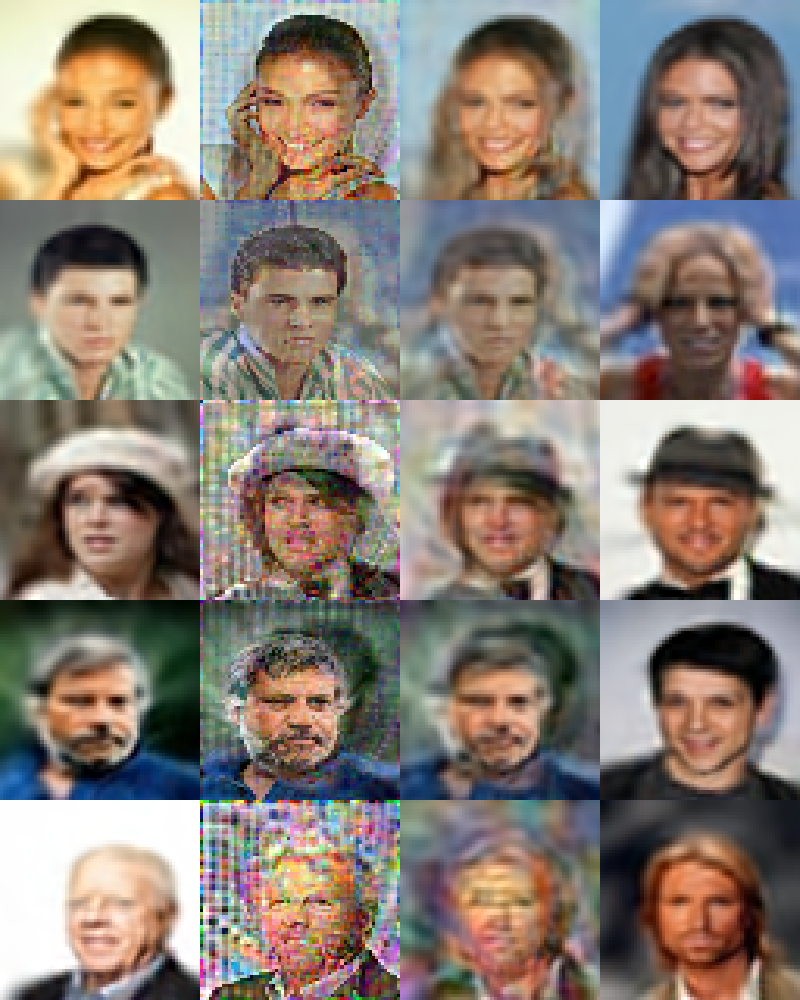}
          \caption{}
      \end{subfigure}
      \begin{subfigure}[b]{0.0525\textwidth}
          \includegraphics[width=\textwidth]{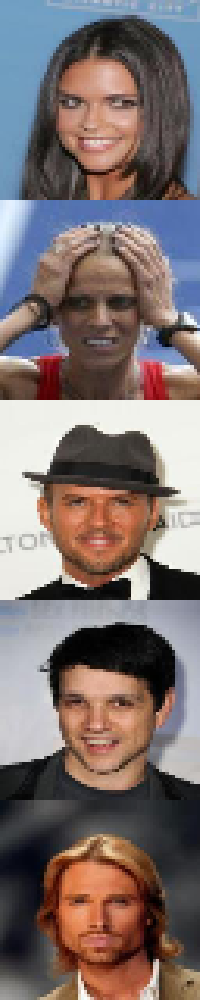}
          \caption{}
      \end{subfigure}
            
      \caption{Visual analysis of results. Original image (a); results for VAE (b), CVAE, (c), DRAW without attention, and (d) DRAW; target image (f). Both DRAW with 16 timesteps. Image-pairs picked at random (from 20 evaluated for each dataset). In each group (b) to (e), from left to right: reconstruction of original (1st); attack input (2nd); attack output (3rd); reconstruction of target (4th). A perfect attack would make (2nd) indistinguishable from (a) and (3rd) indistinguishable from (4th).}
      \label{fig:qualitativeResults}
\end{figure}

The qualitative results appear in Figure~\ref{fig:qualitativeResults}. Although some features are immediately visible, to better appreciate the details, we suggest zooming in the digital version of the article. We contrast VAE, CVAE, DRAW without attention, and DRAW (both with 16 timesteps). We picked the most successful attack for each model (latent layer for VAE and CVAE, output layer for DRAW), the opposite case is available as supplemental material. For each dataset, we sampled at random five image pairs from the twenty used in the evaluation. For each experiment, a single image-pair consists of 51 attacks (different values of the regularization constant). In each case, we chose a mid-way attack, the closest to the average in the horizontal axis of the Distortion--Distortion plot (as shown in the red dot in Figure~\ref{fig:metric})

Attacking autoencoders is clearly difficult: no attack succeed in reconstructing the target image well without incurring in immediately visible distortions to the input. Again, the superior resistance of DRAW with attention appears: the attacks fail to reach the target, incur in large distortions to the input, or both. Although we only attempted targeted attacks, untargeted resistance can be appreciated to some extent, by comparing in each group (b) to (e) the 1st and the 3rd columns: if the model resisted the attack, those columns should be nearly identical.

\section{Discussion}

Attacking autoencoders remains a hard task. No attack can both convincingly reconstruct the target while keeping the distortions on the input imperceptible. Still, not all attempts are equal: some models are significantly more resistant than others. The AUDDC metric allows quantifying that resistance, bypassing the need to establish a clear-cut criterion of success for the attacks. We attempted other metrics (e.g., the slope of a regression on the Distortion--Distortion plots) but found that the AUDDC is better correlated with our subjective perception of resistance to attacks. 

We expected attacks on SVHN to be more challenging than on MNIST, but neither the quantitative or the qualitative analyses confirmed this: the easiest dataset was SVHN, and the harder (as expected) was CelebA. Maybe the smaller surface of attack of MNIST (28$\times$28 input values) compensates for its simplicity, while the huge complexity of CelebA compensates for its larger surface of attack. 
Our results suggest a correlation between the autoencoders intrinsic quality and its resistance to attack. Such correlation does not exist for classifiers, where the best models are not necessarily less susceptible~\cite{tabacof2015exploring}.
Precisely what makes a dataset or model harder to attack is still an exciting open question for future works.

The literature on adversarial attacks on autoencoders is extremely scarce. We expect this to change as autoencoders are advanced as compression schemes for data transmission and storage --- scenarios in which their safety will become paramount.

\subsubsection*{Acknowledgments}
We thank Guilherme de Lázari for initial ideas and code, and Julia Tavares for images and ideas. We thank Brazilian agencies CAPES, CNPq and FAPESP for financial support. We gratefully acknowledge NVIDIA Corporation for donating a Tesla K40 and a Titan X. George Gondim-Ribeiro is funded by FAPESP grant (2017/03706-2) and Eduardo Valle is partially supported by Google Awards LatAm 2017 grant, and by CNPq PQ-2 grant (311905/2017-0).

\FloatBarrier 

\bibliographystyle{unsrt}
{\footnotesize
\bibliography{refs.bib}}

\end{document}